\documentclass[10pt,twocolumn,letterpaper]{article}

\usepackage{cvpr}
\usepackage{times}
\usepackage{epsfig}
\usepackage{graphicx}
\usepackage{amsmath}
\usepackage{amssymb}
\usepackage{tabularx}
\usepackage{booktabs}

\usepackage[breaklinks=true,bookmarks=false]{hyperref} 

\cvprfinalcopy 

\def\cvprPaperID{1203} 

\newcommand{\argmax}{\operatornamewithlimits{arg\,max}}

\def\cf{\emph{cf}\onedot}

\ifcvprfinal\fi
\begin{document}

\title{NeuralNetwork-Viterbi: A Framework for Weakly Supervised Video Learning}

\author{Alexander Richard, Hilde Kuehne, Ahsan Iqbal, Juergen Gall\\
University of Bonn, Germany\\
{\tt\small \{richard,kuehne,iqbalm,gall\}@iai.uni-bonn.de}
}

\maketitle

\begin{abstract}
    Video learning is an important task in computer vision and has experienced
    increasing interest over the recent years. Since even a small amount of videos
    easily comprises several million frames, methods that do not rely on a frame-level
    annotation are of special importance.
    In this work, we propose a novel learning algorithm with a
    Viterbi-based loss that allows for online and incremental learning of weakly
    annotated video data. We moreover show that explicit context and length modeling
    leads to huge improvements in video segmentation and labeling tasks and
    include these models into our framework.
    On several action segmentation benchmarks, we obtain an improvement of up to
    10\% compared to current state-of-the-art methods.
\end{abstract}
\vspace{-0.5cm}

\section{Introduction}
\label{sec:introduction}

A continuously growing amount of publicly available video data on YouTube
or video streaming services, an increased interest in applications
such as surveillance, and the need to analyze continuous video streams \eg in the
domain of autonomous driving has caused an increased interest in video learning
algorithms.

While approaches for action classification on pre-segmented video clips already
perform convincingly well~\cite{simonyan2014two, wang2016temporal,carreira2017quovadis,feichtenhofer2017residual},
realistic applications require the segmentation of temporally untrimmed videos
that usually contain a large variety of different actions with different lengths.
Since acquiring frame-level annotations of such videos is expensive,
methods that can learn from less supervision are of particular interest.
A popular type of weak supervision are transcripts~\cite{bojanowski14weakly,huang2016connectionist,kuehne2017weakly,richard2017weakly,koller2016deephand,koller2017resign}, which provide for each training video an ordered list of actions, but not the frames where the actions occur in the video.


In order to learn a model for temporal action segmentation with such weak supervision, CNNs or RNNs have been combined 
with an explicit model for the intra-class temporal progression, \eg a hidden Markov model (HMM), and the
inter-class context, \eg with a finite grammar~\cite{richard2017weakly,koller2016deephand,koller2017resign}. While these approaches are particularly suited for videos that contain complex
actions and have a huge number of distinct classes, they come with the major problem
that their training requires some heuristical ground truth. They rely on a two-step approach that is iterated several times. It consists of first generating a segmentation for each training video using the Viterbi algorithm and then training the neural network as in the fully supervised case using the generated segmentation as pseudo ground-truth. Consequently, the two-step approach is sensitive to the initialization of the pseudo ground-truth and the accuracy tends to oscillate between the iterations~\cite{richard2017weakly}. In contrast to theses methods, CTC~\cite{graves2006connectionist} is a framework for weakly supervised sequence learning. However, this approach does not allow to include explicit models for the context between classes and their temporal progression and therefore does not achieve state-of-the-art performance.

\begin{figure}
    \centering
    \includegraphics[scale=0.8]{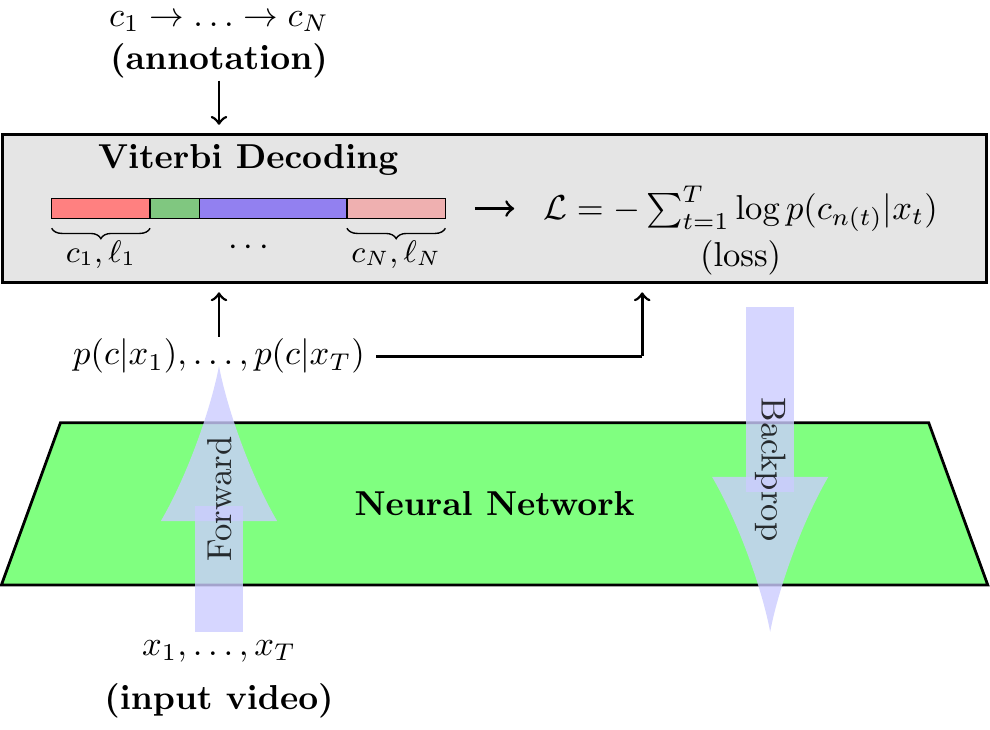}
    \caption{The input video $ \mathbf{x}_1^T $ is forwarded through the network and the
             Viterbi decoding is run on the output probabilities. The frame labels generated
             by the Viterbi algorithm are then used to compute a framewise cross-entropy loss
             based on which the network gradient is computed.}
    \label{fig:onlineViterbi}
\end{figure}

In this work, we propose a novel learning algorithm that allows for direct
learning using the input video and ordered action classes only. The approach includes the
Viterbi-decoding as part of the loss function to train the neural network and has several
practical advantages compared to the two-stage approach:
it neither suffers from an oscillation effect nor requires a frame-wise labeling as
initialization or any kind of pseudo ground-truth, models are learned incrementally,
and the accuracy is improved due to direct optimization of the loss function of interest.

As a second contribution, we also propose to use an explicit length model instead of the widely used
HMMs~\cite{kuehne2017weakly,richard2017weakly,koller2016deephand,koller2017resign},
allowing to learn the action classes directly rather than intermediate HMM states.
In an extensive evaluation, we show an increase of up to $ 10\% $ in accuracy
compared to existing methods.

\section{Related Work}
\label{sec:related_work}

Most existing work on temporal action segmentation focuses on a fully supervised
task~\cite{kuehne16end,rohrbach2012database,tang2012learning,lea2016segmental,vo2014stochastic,ni2014multiple,eyjolfsdottir2014detecting,zhao2017segment,lea2017temporal,yeung2016endtoend,singh16multistream}.
Purely CNN based approaches such as structured segment networks~\cite{zhao2017segment}
or temporal convolutional networks~\cite{lea2017temporal} have recently shown
convincing results on several action segmentation benchmarks. Similarly, LSTM-based
approaches have been in focus~\cite{yeung2016endtoend,singh16multistream}. However,
formulated in a classical deep learning setting, these approaches all rely on fully
supervised, \ie framewise annotated training data.
Richard and Gall~\cite{richard2016temporal} propose
the use of an explicit statistical language model as well as a length model in a fully
supervised formulation.
Note that a frame-level annotation is available, so the estimation of the length model is straightforward
using the ground-truth length from the training set. Moreover, the length model is
learned prior to the actual action classifier. In our approach, on the contrary, no
frame-level annotation is provided, so the length model changes over time during the
training process and is dependent on all other models.


First attempts on weakly supervised learning have been made by~\cite{laptev08learning,marszalek09actions},
who try to to obtain training examples based on movie scripts.
Duchenne \etal~\cite{duchenne09automatic} first addressed the problem of segmenting
actions within videos, assuming that a clip contains not only frames of an action
but also background frames.
Current approaches go much further and try to infer an exact temporal segmentation
of multiple action classes within a single video using weakly annotated training
data only.
In~\cite{sun15temporal,gan2016webly,gan2016youlead}, web images are used to guide video learning in a weakly supervised fashion.
Wang \etal~\cite{wang2017untrimmed} use a purely CNN-based approach to weakly
supervised action detection, where they use a different type of supervision, namely 
short unordered action lists. This approach is well suited to detect action occurrences in a video with large
background portions but not designed for videos that contain a huge amount of
different action classes as in our case.
A similar kind of supervision using unordered sets of actions is proposed
in~\cite{richard2018actionsets}, who also rely on the model factorization proposed by~\cite{richard2016temporal}.
In~\cite{bojanowski14weakly}, the task of temporal action segmentation is relaxed to an
alignment task: it is assumed that an ordered list of occurring actions is also
given for inference, thus it only remains to align those action classes to the video frames.

Kuehne \etal~\cite{kuehne2017weakly} interpret the task of learning an
action segmentation system just given an ordered list of occurring actions as supervision
as an instance of a speech recognition problem, where the videos correspond to the audio
signal and the action classes correspond to words. They apply a standard HMM-GMM system
using a speech recognition toolkit.
Building upon this idea, Richard \etal~\cite{richard2017weakly} replace the GMM by
a recurrent neural network but still rely on an HMM for a coarse temporal modeling. 
A similar approach has been proposed by Koller~\etal~\cite{koller2016deephand,koller2017resign}
for sign language recognition, which is a problem closely related to temporal action segmentation,
but with the most significant difference that the actual temporal boundaries of the recognized words/classes are not relevant.
Note that in contrast to our proposed method, \cite{kuehne2017weakly,richard2017weakly,koller2016deephand,koller2017resign}
use a two-step optimization scheme that does not allow for direct, sequence-wise training.

Lin \etal~\cite{lin2017ctc} use the CTC approach in combination with a statistical language model
for weakly supervised video learning. However, they only infer the sequence of actions occurring in the video.
As an extension of the CTC approach, \cite{huang2016connectionist} propose ECTC
that takes visual similarities between the frames into account to avoid degenerate
segmentations. In contrast to our method, this approach does not allow to include explicit context
and length models.

\section{Temporal Action Segmentation}

We address the problem of temporally localizing activities in a video $\mathbf{x}_1^T = (x_1,\dots,x_T)$ with $ T $ frames.  
The task is to find a segmentation of a video  into an unknown number of $ N $ segments and to
output class labels $ \mathbf{c}_1^N = (c_1,\dots,c_N) $ and lengths $ \mathbf{l}_1^N = (\ell_1,\dots,\ell_N) $
for each of the $ N $ segments. Using a background class for uninteresting frames, each frame
can be assigned to a segment.
For terms of simplicity, we refer to the label assigned to frame $ x_t $ as $ c_{n(t)} $, where
$ n(t) $ is the number of the segment frame $ t $ belongs to.
%
Putting the task in a probabilistic setting, we aim to find the most likely video
labeling given the video frames, \ie
\begin{align}
    (\mathbf{\hat c}_1^N, \mathbf{\hat l}_1^N) = \argmax_{\mathbf{c}_1^N, \mathbf{l}_1^N} \big\{ p(\mathbf{c}_1^N, \mathbf{l}_1^N | \mathbf{x}_1^T) \big\}.
    \label{eq:generalModel}
\end{align}

State-of-the-art methods~\cite{vo2014stochastic,richard2016temporal,kuehne2017weakly,koller2016deephand,richard2017weakly,koller2017resign} formulate $p(\mathbf{c}_1^N, \mathbf{l}_1^N | \mathbf{x}_1^T)$ in such a way such that the $ \argmax $ can be efficiently computed using a Viterbi-like algorithm. Depending on the approach, the models are either trained in a fully supervised setting~\cite{vo2014stochastic,richard2016temporal}, which requires a very time-consuming frame-wise labeling of the training videos, or in a weakly supervised setting~\cite{kuehne2017weakly,koller2016deephand,richard2017weakly,koller2017resign}. In the latter case, the training videos are annotated only by an ordered sequence of action classes that occur in the video. This means each training instance is a tuple $ (\mathbf{x}_1^T, \mathbf{c}_1^N) $ consisting of a video $ \mathbf{x}_1^T $ and a transcript sequence $ c_1 \rightarrow \dots \rightarrow c_N $ that defines the ordering
of occurring actions. In contrast to the fully supervised setting, $\mathbf{l}_1^N$ and accordingly the frame-level annotation of the training data is unknown.

In this work, we focus on the problem of weakly supervised learning and propose two contributions. The first contribution addresses the modeling of $p(\mathbf{c}_1^N, \mathbf{l}_1^N | \mathbf{x}_1^T)$. Instead of using a hidden Markov model as in \cite{kuehne2017weakly,koller2016deephand,richard2017weakly,koller2017resign}, we explicitly model the length of each action class. The model is described in Section~\ref{sec:model} and in our experiments we show that the proposed length model outperforms an HMM. The second contribution is a more principled approach for weakly supervised learning. This approach is described in Section~\ref{sec:end2endViterbi} and can be used to train any model that uses neural networks and Viterbi decoding such as~\cite{koller2016deephand,richard2017weakly,koller2017resign}.


\section{NeuralNetwork-Viterbi}
\label{sec:end2endViterbi}

Before we describe the proposed learning approach in Section~\ref{sec:Viterbi}, we briefly summarize the training in a fully supervised setting and the training procedure that is used in~\cite{kuehne2017weakly,koller2016deephand,richard2017weakly,koller2017resign} for weakly supervised learning.   
 
In a classical fully supervised training setup, frame-wise ground-truth annotation
is provided for the training data, \ie each training video comprises the triple $ (\mathbf{x}_1^T, \mathbf{c}_1^N, \mathbf{l}_1^N) $. 
Since $\mathbf{l}_1^N$ and therefore the label $ c_{n(t)} $ for each frame $ x_t $ is known, the underlying model for Equation~\eqref{eq:generalModel}, which is typically a neural network, is trained using the frame-level annotations and, for instance, the cross-entropy loss.


If only the transcript of a training video, \ie an ordered sequence of classes that occur in the video, is given, $\mathbf{l}_1^N$ is unknown and only $ (\mathbf{x}_1^T, \mathbf{c}_1^N) $ is provided. Most existing weakly supervised approaches~\cite{kuehne2017weakly,koller2016deephand,richard2017weakly,koller2017resign}
reduce the problem to the fully supervised case by generating a pseudo ground-truth $c^{\text{pseudo}}_{n(t)}$
for all training sequences. A neural network is then trained using a pseudo cross-entropy
loss that is based on the pseudo ground-truth $c^{\text{pseudo}}_{n(t)}$. 

This approach comes with a major problem:
The model learning and transcript decoding (\ie pseudo ground-truth generation) are
separated and the transcripts $ \mathbf{c}_1^N $ are only used for the pseudo ground-truth $c^{\text{pseudo}}_{n(t)}$
generation. In other words, the model learning does not explicitly include the transcripts.
As a workaround, the two steps \textit{pseudo ground-truth generation} and \textit{model learning}
are repeated several times, where the pseudo ground-truth in the first iteration
is a uniform alignment of transcripts to sequence frames.
In later repetitions, the pseudo ground-truth is generated using a Viterbi decoding on Equation~\eqref{eq:generalModel} with the previously trained network. From a practical point, this results in several major limitations. As it was reported in~\cite{richard2017weakly}, the approach is sensitive to the initialization of the pseudo ground-truth and the accuracy tends to oscillate between the iterations. Furthermore, the approach processes in each step the entire dataset, which prevents its use for incremental learning.              

In this work, we propose a new framework that allows to learn directly from the
transcripts. Therefore, we define a loss that can be computed solely based on the
current model and a single training example $ (\mathbf{x}_1^N, \mathbf{c}_1^N) $.
The loss is designed to be zero if
\begin{align}
    p(\mathbf{c}_1^N, \mathbf{l}_1^N | \mathbf{x}_1^T) = p(\mathbf{c}_1^N, \mathbf{l}_1^N | \mathbf{x}_1^T, \mathbf{c}_1^N),
\end{align}
\ie if the prediction without given transcripts (left hand side) is equal to the prediction
with given transcripts (right hand side).
Particularly, our approach does not require a precomputed pseudo ground-truth and works
directly on the weakly annotated data.

\subsection{Viterbi-based Network Training}
\label{sec:Viterbi}

Our new training procedure is illustrated in Figure~\ref{fig:onlineViterbi}.
The training algorithm randomly draws a sequence $ \mathbf{x}_1^T $
and its annotation $ \mathbf{c}_1^N $ from the training set. The sequence is then
forwarded through a neural network. Note that there are no constraints on the network
architecture, all commonly used feed-forward networks, CNNs, and recurrent networks
can be used.
The optimal segmentation by means of Equation~\eqref{eq:generalModel} is then
computed by application of a Viterbi decoding on the network output, see Section~\ref{sec:viterbiRevisited}
for details. Since $ \mathbf{c}_1^N $ is provided as annotation, only $ \mathbf{l}_1^N $ needs
to be inferred during training.
We switch notation and write the Viterbi segmentation $ (\mathbf{c}_1^N, \mathbf{l}_1^N) $
as framewise labels $ c_{n(1)}, \dots, c_{n(T)} $, with which the cross-entropy loss over all aligned frames
is accumulated:
\begin{align}
    \mathcal{L} = -\sum_{t=1}^T \log p(c_{n(t)}|x_t).
\end{align}
We chose the cross-entropy loss as it is most common in neural network
optimization. However, our framework is not bound to a specific loss function.
Once the Viterbi segmentation of the input sequence is computed, any other
loss such as squared-error can as well be used.

Based on the sequence loss $ \mathcal{L} $, the network parameters are updated
using stochastic gradient descent with the gradient $ \nabla \mathcal{L} $ of the
loss. We would like to emphasize that the algorithm operates in an online fashion,
\ie in each iteration, the loss $ \mathcal{L} $ is computed with respect to a single
randomly drawn training sequence $ (\mathbf{x}_1^T, \mathbf{c}_1^N) $ only.

\subsection{Enhancing the Robustness}
\label{sec:robust}

In practice, a sequence $ \mathbf{x}_1^T $ can easily be a few thousand frames long.
Backpropagating all frames at once can thus raise problems with the limited GPU memory.
Moreover, online learning algorithms generally benefit from making a large number of model
updates. Therefore, we split the sequence into multiple mini-batches after the Viterbi segmentation
$ c_{n(1)},\dots,c_{n(T)} $ has been computed. These minibatches are then backpropagated
one-by-one through the network.

However, traditional online learning algorithms such as stochastic gradient descent rely on
the assumption that
\begin{align}
    \mathcal{L}^*(w) = \mathbb{E}_{\mathbf{x}} \mathcal{L}(x, w)
                     = \int \mathcal{L}(x, w) d\mathcal{P}(x),
\end{align}
where $ w $ denotes the model parameters, $ \mathcal{L}^*(w) $ is the true loss
that is to be optimized, and $ \mathcal{L}(x,w) $ is the loss of a single observation $ x $,
see \eg~\cite{bottou1998online}.
In each iteration, the observations $ x $ are usually assumed to be drawn independently
from the distribution $ \mathcal{P}(x) $. In our setting, on the contrary, all frames in
an iteration belong to the same sequence $ \mathbf{x}_1^T $, so they are
not independent. Further subdividing long sequences into smaller mini-batches
enhances the problem: multiple updates are made with a strong bias towards \textit{(a)}
the characteristics of the sequence frames and \textit{(b)} the limited amount of
classes occurring in the sequence.

We therefore propose to use a buffer $ \mathcal{B} $ and store recently processed sequences
and their inferred frame labels. In order to make the gradient in each iteration
more robust, $ K $ frames from the buffer are sampled and added to the loss function,
\begin{align}\label{eq:buffer}
    \mathcal{L} = -\Big[ \sum_{t=1}^T \log p(c_{n(t)}|x_t) + \sum_{k=1}^K \log p(c_k|x_k) \Big].
\end{align}
Since the neural network is updated gradually in small steps, most of the frame/label pairs
in the buffer still agree with the current model. However, sampling random frames from
the buffer lessens the above-mentioned sequence bias from the loss function and
increases the robustness of the optimization algorithm.

\section{The Model}
\label{sec:model}

We now introduce the specific model used in this paper. Starting from Equation~\eqref{eq:generalModel},
we factorize the overall probability $ p(\mathbf{c}_1^N, \mathbf{l}_1^N | \mathbf{x}_1^T) $,
\begin{align}
    (\mathbf{\hat c}_1^N, \mathbf{\hat l}_1^N) &= \argmax_{\mathbf{c}_1^N, \mathbf{l}_1^N} \big\{ p(\mathbf{c}_1^N, \mathbf{l}_1^N | \mathbf{x}_1^T) \big\} \nonumber \\
                                               &= \argmax_{\mathbf{c}_1^N, \mathbf{l}_1^N} \big\{ p(\mathbf{x}_1^T|\mathbf{c}_1^N,\mathbf{l}_1^N) \cdot p(\mathbf{l}_1^N|\mathbf{c}_1^N) \cdot p(\mathbf{c}_1^N) \big\}.
\end{align}
Assuming conditional independence of the frames, the $ \argmax $ term can be further decomposed into
\begin{align}
    \argmax_{\mathbf{c}_1^N, \mathbf{l}_1^N} \Big\{ \prod_{t=1}^T p(x_t|c_{n(t)}) \cdot \prod_{n=1}^N p(\ell_n|c_n) \cdot p(c_n|\mathbf{c}_1^{n-1}) \Big\}.
    \label{eq:model}
\end{align}
We refer to $ p(x_t|c_{n(t)}) $ as \textit{visual model}, to $ p(\ell_n|c_n) $ as
\textit{length model}, and to $ p(c_n|\mathbf{c}_1^{n-1}) $ as \textit{context model}.

The \textit{visual model} is a neural network as illustrated in Figure~\ref{fig:onlineViterbi}. We use a recurrent network with a single layer of $ 256 $ gated recurrent units
and a softmax output. Similar recurrent networks have also been used in other
recent methods~\cite{huang2016connectionist,richard2017weakly}, but we train the network as described in Section~\ref{sec:end2endViterbi}. Since the outputs of the neural network are posterior probabilities $ p(c|x_t) $, we follow the hybrid approach~\cite{bourlard2012connectionist} and refactor
\begin{align}
    p(x_t|c) \propto \frac{p(c|x_t)}{p(c)},
    \label{eq:hybrid}
\end{align}
where $ p(c) $ is a class prior. 
During training, we count the amount of frames that have
been labeled with a class $ c $ for all sequences that have been processed so far.
Normalizing these counts to sum up to one finally results in our estimate of $ p(c) $.
The prior is updated after every iteration, \ie after every new training sequence.
If a sequence annotation $ \mathbf{c}_1^N $ contains a class that has not been seen
before, $ 1 / \mathtt{\#classes} $ is used.

As \textit{length model}, we use a class-dependent Poisson distribution:  
\begin{equation}
p(\ell|c) = \frac{\lambda_c^\ell\exp\left(-\lambda_c\right)}{\ell!}. 
\end{equation}
After each iteration, we update $\lambda_c$, which is the mean length of a segment for class $c$.
If the training sample $ (\mathbf{x}_1^T,\mathbf{c}_1^N) $ contains a class that has not been seen before, we set \mbox{$ \lambda_c = N/T $.}

Previous works using \textit{context models} either rely on an
$ n $-gram language model~\cite{richard2016temporal,koller2017resign} or a finite
set of allowed class sequences~\cite{kuehne2017weakly,richard2017weakly}. In order
to capture both possibilities, we use a right-regular stochastic grammar, where all rules
are of the form $ \tilde h \rightarrow c~h $ with $ \tilde h, h $ denoting nonterminal symbols and a class
$ c $ that acts as terminal symbol.
Such a grammar is a superclass of $ n $-grams and finite grammars. 
Therefore, we decode the possible contexts $ \mathbf{c}_1^{n-1} $ as non-terminal
symbols $ h $ of the grammar and denote the probability to hypothesize class $ c $
given the context $ h $ as $ p(c|h) $. During training, the grammar for a sequence is defined by the transcript sequence $\mathbf{c}_1^{N}$. For evaluation, we consider two tasks, namely action alignment and action segmentation. While for action alignment a transcript sequence, which defines the grammar, is also provided for each test sequence, transcripts are not provided for the more difficult task of action segmentation. In this case, we estimate the grammar from the transcript annotation of all training videos.


\subsection{Viterbi Algorithm Revisited}
\label{sec:viterbiRevisited}

Finding the best segmentation in terms of Equation~\eqref{eq:model} is a challenging
problem given the exponentially large search space over all possible class sequences
and lengths. Most works optimizing a similar quantity rely on the Viterbi algorithm~\cite{kuehne2017weakly,koller2016deephand,koller2017resign,richard2017weakly}
that is based on dynamic programming and is usually used to find the best label sequence
of a hidden Markov model.

In contrast to the standard applications of the Viterbi algorithm, our model additionally
features a length model that makes the optimization more complex. To find the best
sequence by means of Equation~\eqref{eq:model}, we define an auxiliary function
$ Q(t,\ell,c,h) $ that yields the best probability score for a segmentation up to
frame $ t $ meeting the following conditions:
\begin{enumerate}
    \setlength{\itemsep}{0pt}
    \setlength{\parskip}{0pt}
    \item the length of the last segment is $ \ell $,
    \item the class label of the last segment is $ c $,
    \item the context (the nonterminal symbol) of the stochastic grammar is $ h $.
\end{enumerate}
The function can be computed recursively. We distinguish two cases. The first case
is when no new segment is hypothesized, \ie $ \ell > 1 $. Then,
\begin{align}
    Q(t,\ell,c,h) = Q(t-1,\ell-1,c,h) \cdot p(x_t|c),
\end{align}
so the score of the current frame is multiplied with the auxiliary function's value at the previous frame.
The second case is a new segment being hypothesized at frame $ t $, \ie $ \ell = 1 $. Then,
\begin{align}
    &Q(t,\ell=1,c,h) = \nonumber \\
    &\max_{\substack{ \tilde \ell, \tilde c, \tilde h: \\ \tilde h \rightarrow c~h ~\land \\ \exists h': h' \rightarrow \tilde c ~\tilde h }}
    \big\{ Q(t-1,\tilde \ell,\tilde c,\tilde h) \cdot p(x_t|c) \cdot p(\tilde \ell|\tilde c) \cdot p(c|\tilde h) \big\},
    \label{eq:dynProgBoundaries}
\end{align}
\ie the maximization is carried out over all possible lengths $ \tilde \ell $ and
over all $ \tilde c, \tilde h $ that are a right-hand side of a rule in the grammar
and there is another rule that allows a transition from $ \tilde h $ to $ h $ by
hypothesizing class $ c $.
 
The most likely segmentation of the complete video is then given by
\begin{align}
    \max_{\ell,c,h} \big\{ Q(T,\ell,c,h) \cdot p(\ell|c) \big\}.
\end{align}
The optimal class labels $ \mathbf{c}_1^N $ and lengths $ \mathbf{l}_1^N $ can be
obtained by keeping track of the maximizing arguments $ \tilde c $ and $ \tilde \ell $
from Equation~\eqref{eq:dynProgBoundaries}. Additional details and code are available online.\footnote{\url{https://alexanderrichard.github.io}}

\textbf{Complexity.}
The maximization over $ \ell $ is bounded by the length
$ T $ of the video and the possibilities for $ \tilde c, \tilde h $ pairs are limited
by the number of rules in the grammar, so the cost to compute $ Q $ for a frame $ t $
is linear in the video length and the grammar size. Since $ Q $ needs to be computed for
all frames, the overall complexity is quadratic in the video length. This can
be prohibitive for long videos. Thus, in practice, we limit the maximal allowed length to
a constant $ L $, so the runtime of the Viterbi decoding is linear in both, video length
and grammar size. Throughout this work, we use $ L = 2,000 $.

%
%
%

\section{Experiments}
\label{sec:experiments}


We provide results on three different datasets. The main evaluation (Sections~\ref{sec:robustness} to~\ref{sec:incremental}) is conducted on
\textbf{Breakfast}~\cite{kuehne14language}, a large-scale dataset for action segmentation.
It comprises $ 1,712 $ videos (around $ 3.6 $ million frames) of persons making breakfast.
There are ten dishes such as \textit{pancakes} or \textit{cereals}, all with fine-grained
annotations like $ \textit{stir} $ or $ \textit{pour} $. Overall, there are $ 48 $ action
classes and an average of $ 6.9 $ action instances per video. The videos
range from some seconds to several minutes in length. We follow~\cite{kuehne14language} and
report frame accuracy averaged over four splits.

The \textbf{50 Salads}~\cite{stein2013combining} dataset is another video dataset for action segmentation. Although
it only contains $ 50 $ videos, each video is very long and the dataset still has nearly
$ 600,000 $ frames annotated with $ 17 $ classes, which amount to an average of $ 20 $
action instances per video. As evaluation metric, we report frame accuracy averaged
over five splits.

\textbf{Hollywood Extended} has been introduced in~\cite{bojanowski14weakly} for the
task of action alignment, which we address with our method in Section~\ref{sec:actionAlignment}.
The dataset comprises $ 937 $ videos (nearly $ 800,000 $ frames) and $ 16 $ different classes. Each video
contains $ 2.5 $ action instances on average. As evaluation metric, we follow~\cite{bojanowski14weakly}
and report the Jaccard index as intersection over detection.

\textbf{Setup.}
In accordance with~\cite{kuehne2017weakly,richard2017weakly,huang2016connectionist},
we extract Fisher vectors of improved dense trajectories~\cite{wang2013action}
over a temporal window of length $ 20 $ for each frame and reduce the result to
$ 64 $ dimensions using PCA.
In all experiments, the recurrent network is trained for $ 10,000 $ iterations with a learning rate of $ 0.01 $
for the first $ 2,500 $ iterations and $ 0.001 $ afterwards.
The minibatch size for backpropagation of the frames of a training sequence (\cf Section~\ref{sec:robust})
is set to $ 512 $.


\subsection{Robustness}
\label{sec:robustness}

We start with an evaluation of our proposed end-to-end learning algorithm. As discussed in Section~\ref{sec:robust}, we enhance the loss function~\eqref{eq:buffer} by sampling additional frames from a buffer. In the following, we evaluate the impact of this enhancement and its parameters, namely number of sampled frames and buffer size.  


\textbf{Impact of Old Data Sampling.}
The first proposition to enhance the robustness of our algorithm is to maintain
some recently seen sequences and their inferred labeling in a buffer and to sample
a certain amount $ K $ of additional frames from this buffer. This way, we want
to ensure that in each iteration, the overall data and class distribution are
sufficiently well captured. For the purpose of analyzing which value for $ K $
is necessary, we assume an unlimited buffer size, \ie all previously processed
sequences are maintained in memory.
\begin{figure}[tb]
    \centering
    \includegraphics{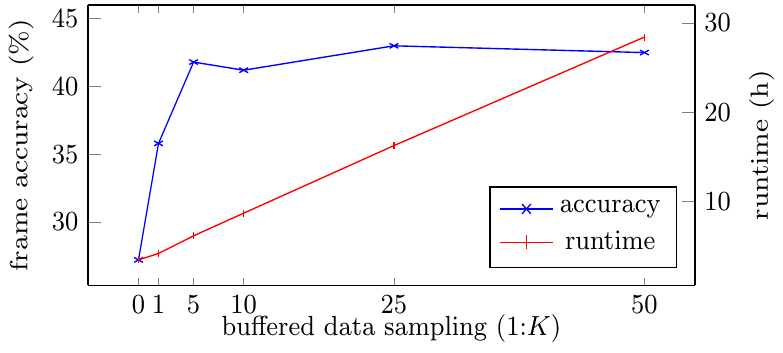}
    \footnotesize
    \begin{tabularx}{0.48\textwidth}{Xrrrrrr}
        \toprule
            ratio $1$:$K$    & $ - $    & 1:1      & 1:5      & 1:10     & 1:25     & 1:50     \\
        \midrule
            accuracy         & $ 27.2 $ & $ 35.8 $ & $ 41.8 $ & $ 41.2 $ & $ 43.0 $ & $ 42.5 $ \\
            runtime (h)      & 3:26     & 4:09     & 6:08     & 8:40     & 16:15   & 28:25     \\
        \bottomrule
    \end{tabularx}
    \caption{\label{fig:bufferedDataSampling}
		Impact of buffered data sampling. A sampling ratio of $1$:$K$ means that for each frame of
             the current sequence, $ K $ buffered frames are sampled.
             The first column shows the result for on-line learning, \ie, without a buffer.
             Runtime is measured on a K80.}
    \vspace{-0.5cm}
\end{figure}
The results are illustrated in Figure~\ref{fig:bufferedDataSampling}.
If we do not sample from previously seen sequences, the model is learned on-line, \ie the training sequences are directly processed and not stored in a buffer. In this case, our approach achieves a frame accuracy of $27.2\%$. If we use a buffer and sample frames from it, the accuracy is greatly increased. 
Without sampling from the buffer, the model learns a strong bias towards the characteristics
and class distributions of the current video only. This can be avoided by adding
other frames from different classes and sequences to the loss function. While a
$1$:$1$ sampling, \ie for each frame in the sequence one buffered frame is sampled,
already shows a huge improvement, we find the optimization to stabilize at a sampling rate
of $1$:$25$. Thus, we stick to this value in all remaining experiments.


\textbf{Impact of the Buffer Size.}
For the above evaluation, we assumed an unlimited buffer size, \ie every processed sequence
could be stored. This may be undesirable in case of large datasets for two reasons:
first, depending on the amount of training data, it can be prohibitive to maintain all videos
in memory at the same time. Second, the underlying assumption when using the buffer is that
the frame/label pairs that are sampled are still more or less consistent with the current
model. While this assumption is reasonable if all buffered sequences have been processed
only a few iterations ago, it will certainly be wrong if there are frame/label pairs that
have been generated by a model a few thousand iterations ago. 
\begin{figure}[tb]
    \centering
    \includegraphics{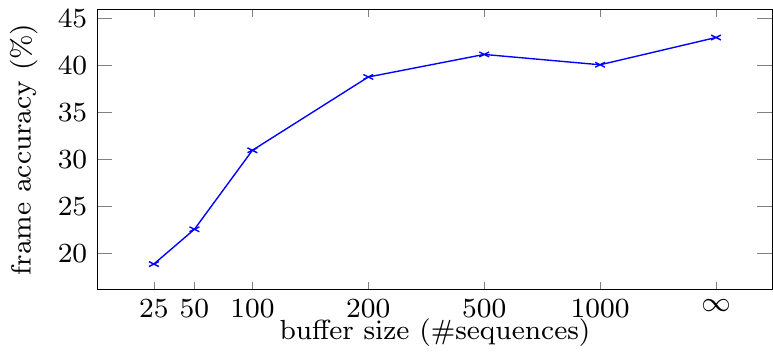}
    \footnotesize
    \begin{tabularx}{0.48\textwidth}{Xr@{\hskip 1.7ex}r@{\hskip 1.7ex}r@{\hskip 1.7ex}r@{\hskip 1.7ex}r@{\hskip 1.7ex}r@{\hskip 1.7ex}r}
        \toprule
            buffer size      & $ 25 $   & $ 50 $   & $ 100 $  & $ 200 $  & $ 500 $  & $ 1000 $ & $ \infty $ \\
        \midrule
            accuracy         & $ 18.9 $ & $ 22.6 $ & $ 31.0 $ & $ 38.8 $ & $ 41.2 $ & $ 40.1 $ & $ 43.0 $   \\
        \bottomrule
    \end{tabularx}
    \caption{Impact of the buffer size for a buffered data sampling ratio of $1$:$25$. Only a few hundred buffered
             sequences are already sufficient for robust learning.}
    \label{fig:bufferSize}
\end{figure}
Hence, we evaluate the impact of the buffer size on the performance, see Figure~\ref{fig:bufferSize}.
Since we already fixed a sampling ratio of $1$:$25$, a buffer size of less than $ 25 $ sequences
is not reasonable. A too small buffer of less than $ 100 $ sequences does
not reflect the overall data and class distributions well enough, resulting in a poor segmentation
performance, \cf~Figure~\ref{fig:bufferSize}.
With more than $ 200 $ buffered sequences, however, the system stabilizes.
Considering the size of the datasets we use (less than $ 2,000 $ sequences each), old frame/label
pairs being inconsistent with the current model are not an issue here. Hence, we leave the buffer size unlimited for the remainder of this work.

\textbf{Batch Size.}
\begin{figure}[tb]
    \centering
    \includegraphics{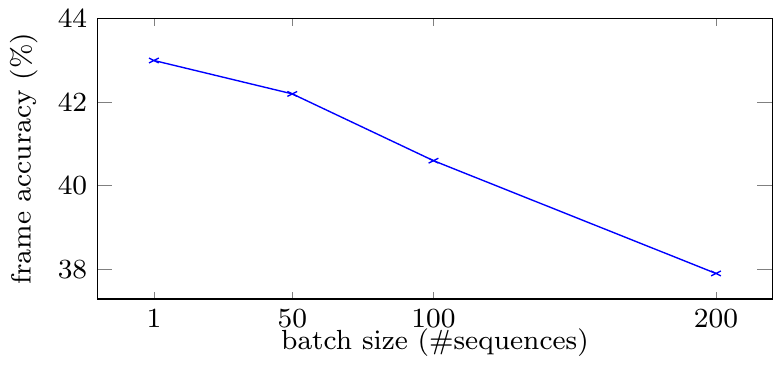}
    \caption{Effect of the batch size. A small batch and frequent updates are beneficial for better accuracy.}
    \label{fig:batchSize}
\end{figure}
In all experiments, we use a batch size of one. Figure~\ref{fig:batchSize} shows that with larger batch
sizes the accuracy slowly drops. Our model is continuously updated, \ie segmentation information from
previous iterations enters the parameter updates, via a running length and prior estimate as well as through
buffered data. Thus, a small batch size allows for a rapid adaptation of the length model and prior.

\textbf{Convergence Behaviour.}
\begin{figure}[tb]
    \centering
    \includegraphics{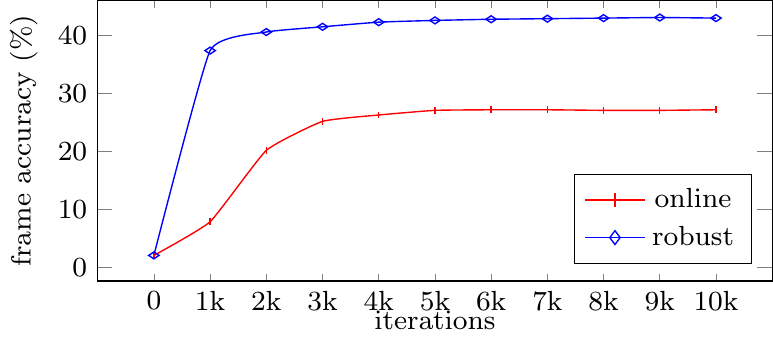}
    \caption{Convergence behaviour of our NN-Viterbi algorithm in both
             variants, online (red) and with enhanced robustness (blue), over $ 10,000 $ training iterations.}
    \label{fig:convergence}
    \vspace{-0.5cm}
\end{figure}
Figure~\ref{fig:convergence} shows the convergence behaviour of our algorithm
as a pure online learning approach (no buffered data sampling) and with the robustness enhancements,
\ie with a $1$:$25$ data sampling and an unlimited buffer size.
While both variants of our algorithm start to converge after $ 2,000 $ to $ 3,000 $
iterations, the robustness enhancement is particularly advantageous at the beginning
of training, adding a huge margin in terms of frame accuracy compared to the pure
online variant. Note that~\cite{richard2017weakly} report an oscillating accuracy
over the iterations of their two-step scheme. Our NN-Viterbi, in contrast, has a
smooth and stable convergence behaviour for both variants.

\subsection{Impact of Direct Learning and Model}
\label{sec:ablation}

\begin{table}
    \footnotesize
    \begin{tabularx}{0.48\textwidth}{Xrr}
        \toprule
                                                               & accuracy (\%) & runtime (h) \\
        \midrule
            pseudo ground-truth~\cite{richard2017weakly}       & $ 23.9 $ & 03:45 \\
            pseudo gr.-tr. + HMM~\cite{richard2017weakly}      & $ 33.3 $ & 08:12 \\
            pseudo gr.-tr. + HMM + LM                          & $ 36.4 $ & 17:21 \\
            pseudo gr.-tr. + LM                                & $ 39.1 $ & 06:04 \\
            NN-Viterbi + LM                                    & $ 43.0 $ & 22:43 \\
        \bottomrule
    \end{tabularx}
    \caption{Impact of length modeling in combination with NN-Viterbi compared
             to different models using a pseudo ground-truth. Training time is measured on a TitanX.}
    \label{tab:ablation}
    \vspace{-0.5cm}
\end{table}

In this section, we evaluate the impact of our proposed algorithm compared
to the state-of-the-art methods for weakly supervised learning which generate pseudo ground-truth instead of using the transcript annotations directly for learning as discussed in Section~\ref{sec:end2endViterbi}, and the advantages of temporal modeling
using an explicit length model rather than an HMM as discussed in Section~\ref{sec:model}.
The results are shown in Table~\ref{tab:ablation}.

\subsubsection{Temporal Modeling: HMM vs.\ Length Model}
Since most recent methods use a hidden Markov model for the temporal progression throughout
the sequence~\cite{kuehne2017weakly,richard2017weakly,koller2017resign}, we first show the
benefits of modeling the temporal progression directly with a length model. Although the
Viterbi decoding is more involved in this case, it allows to train a model directly on the
action classes rather than on hidden Markov model states.
First, note the impact of temporal modeling in general: if we neither use an HMM nor
an explicit length model, the accuracy drastically drops, see first row of Table~\ref{tab:ablation}.
When introducing an HMM as in~\cite{richard2017weakly}, nearly $ +10\% $ improvement can be
observed. Using our factorization from Equation~\eqref{eq:model} with the explicit length
model, however, a further gain of $ +6\% $ is achieved, see fourth row of Table~\ref{tab:ablation}.
The reason for the latter is twofold: First, the training data is aligned to the actual classes
rather than to a huge number of HMM states, so for each class more training examples are available.
Second, the number of HMM states is fix during network training, while the length model can
dynamically adopt to the learned model during training. Notably, using a length model on
HMM states is not recommendable since HMM states are typically of very short duration and the
state-wise length model has no major impact.

\subsubsection{Pseudo Ground-Truth vs.\ Direct Learning}
\begin{figure}[tb]
    \centering
    \includegraphics{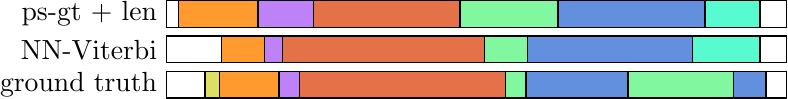}

    \vspace{0.3cm}
    \includegraphics{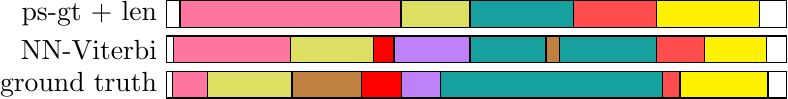}
    \caption{Example segmentations of two videos from the Breakfast dataset. The two-step scheme
             with pseudo ground truth and length model has a bias towards uniform lengths, which prevents short
             actions from being detected accurately. The NN-Viterbi approach is much more robust.}
    \label{fig:qualitative}
    \vspace{-0.3cm}
\end{figure}
Note that so far, the model is still trained according to the two-step paradigm of repeatedly
generating a pseudo ground-truth and optimizing the network. Using our proposed algorithm,
on the contrary, leads to much better results of $ 43.0\% $ accuracy, which can be attributed
to the direct loss, see Table~\ref{tab:ablation}. In case of the two-step scheme, the model is encouraged to learn
the errors that are present in the generated pseudo ground-truth. Including the transcripts
directly into the model learning, this can be avoided.

In Figure~\ref{fig:qualitative}, two example segmentations are shown. Recall that for the two-step
scheme, the initial pseudo ground-truth is a uniform segmentation. Even after several iterations,
a bias towards uniform sequence lengths can be observed. This leads to inaccurate detections of
short segments (upper example segmentation) or even completely missed segments (lower example segmentation).
Our proposed NN-Viterbi learning is much more accurate, specifically when the segment lengths
vary strongly.

\subsection{Incremental Learning}
\label{sec:incremental}

In a classical learning setup, usually a fixed training set is provided. In this case,
it is convenient to process all data in random order.
For algorithms working in an online or incremental fashion,
however, an interesting practical question is what happens if not all training data is
available right at the beginning. For instance, video data from different actors is
added to the training data over time. Or, training data for some classes is only available
at a later point in time.

\begin{figure}
    \centering
    \includegraphics[width=0.48\textwidth]{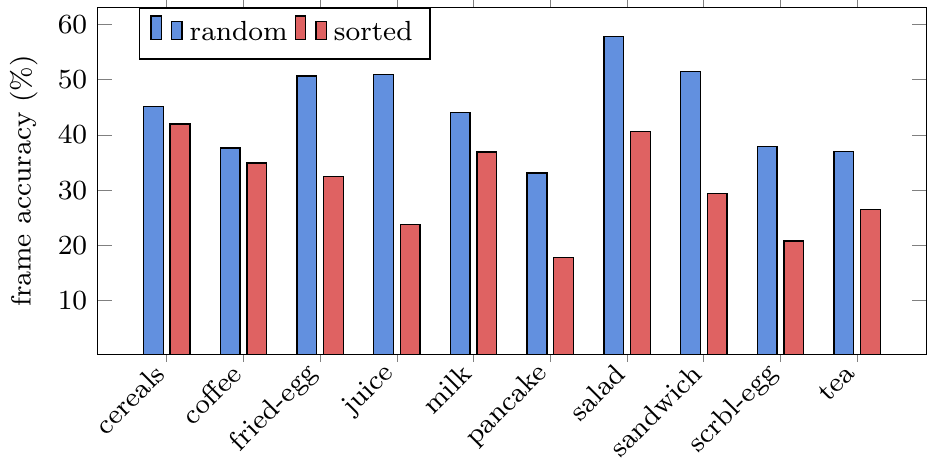}
    \caption{Accuracy per coarse activity for randomly shuffled training data and training data sorted
             by coarse activities. Left activities have been seen early during training, right activities later.}
    \label{fig:sortedActivities}
    \vspace{-0.3cm}
\end{figure}

We therefore analyze our algorithm under such conditions. 
To this end, we sorted the training set \textit{(a)} by the ten coarse Breakfast activities\footnote{Each video in the Breakfast dataset belongs to one of ten coarse activities. The activities are compositions of $ 48 $ fine-grained action classes.}
and \textit{(b)} by the actors, see Table~\ref{tab:inputOrder}.
In the first case, coarse activities that have been observed in the beginning, \eg \textit{cereals} and \textit{coffee},
hardly lose any accuracy compared to training with randomly shuffled data, see Figure~\ref{fig:sortedActivities}.
Later coarse activities are usually not learned well and experience a relative drop of about $ 50\% $ compared to random shuffling.
The comparably small performance drop for \textit{milk} and \textit{tea} is due to the fact that
these activities share a lot of fine-grained action classes with \textit{cereals} and \textit{coffee},
for instance \texttt{take\_cup} or \texttt{pour\_milk}.

\begin{table}
    \footnotesize
    \begin{tabularx}{0.48\textwidth}{Xr}
        \toprule
                                & frame accuracy (\%) \\
        \midrule
            sorted by activity      & $ 27.9 $ \\
            sorted by actor         & $ 41.5 $ \\
            randomly shuffled       & $ 43.0 $ \\
        \bottomrule
    \end{tabularx}
    \caption{Impact of the sequence input order on the robustness of the algorithm.
             The videos are sorted \textit{(a)} by the ten coarse activities of the
             Breakfast dataset, \textit{(b)} by the performing actor, and \textit{(c)}
             randomly.}
    \label{tab:inputOrder}
\end{table}

Compared to the case where all data is available right at the beginning and random shuffling is
possible, sorting the data by actor still results in a very good performance. Apparently,
learning the correct class distributions right at the beginning is very important, while changes in appearance over time - such as changing actors - still allows to robustly learn the underlying concepts of the classes.

\subsection{Comparison to State of the Art}

In this section, we compare our approach to state-of-the-art methods for the same task,
see Table~\ref{tab:actionSegmentation}.
While OCDC~\cite{bojanowski14weakly} is based on a discriminative clustering,
\cite{kuehne16end} and~\cite{richard2017weakly} rely on hidden Markov models and train
their systems with the classical repeated two-step scheme. Their model formulation is
comparable to our factorization from Equation~\eqref{eq:model}. Still, NN-Viterbi outperforms
the methods by a large margin.
CTC and ECTC allow to optimize the posterior probabilities $ p(\mathbf{c}_1^N|\mathbf{x}_1^T) $ directly.
However, the criterion does not include explicit models such as a stochastic
grammar or a length model. The assumption is that the underlying recurrent network can learn
all temporal dependencies on its own. As also shown in~\cite{huang2016connectionist}, this can
lead to degenerate segmentations particularly when videos are long, since even LSTMs usually struggle
to memorize context over multiple hundred frames.
Human actions typically are rather long, hence modeling context and length explicitly is very
important and purely CTC based methods struggle to achieve comparable performance.
Lin \etal~\cite{lin2017ctc} also use a CTC based model on Breakfast to infer the sequence of actions in a video.
They evaluate the unit accuracy, \ie the edit distance between the inferred action transcript
and the ground truth transcript, and obtain $ 43.4\% $ unit accuracy. With our approach,
we obtain $ 55.5\% $ unit accuracy.

\begin{table}
    \footnotesize
    \begin{tabularx}{0.48\textwidth}{Xrr}
        \toprule
                                            & \textbf{Breakfast} & \textbf{50 Salads} \\
        \midrule
        OCDC \cite{bojanowski14weakly}      & $ 8.9 $            & $ - $              \\
        CTC  \cite{huang2016connectionist}  & $ 21.8 $           & $ 11.9 $           \\
        HTK  \cite{kuehne16end}             & $ 25.9 $           & $ 24.7 $           \\
        ECTC \cite{huang2016connectionist}  & $ 27.7 $           & $ - $              \\
        HMM/RNN \cite{richard2017weakly}    & $ 33.3 $           & $ 45.5 $           \\
        \midrule
        \textbf{NN-Viterbi}            & $ \mathbf{43.0} $  & $ \mathbf{49.4} $  \\
        \bottomrule
    \end{tabularx}
    \caption{Comparison of our method to several state-of-the-art methods for the task of
             temporal action segmentation. Results are reported as frame accuracy (\%).}
    \label{tab:actionSegmentation}
\end{table}

\subsection{Action Alignment}
\label{sec:actionAlignment}

\begin{table}
    \footnotesize
    \begin{tabularx}{0.48\textwidth}{Xrr}
        \toprule
                                            & \textbf{Hollywood Extended} \\
        \midrule
        ECTC \cite{huang2016connectionist}  & $ 41.0 $ \\
        HTK \cite{kuehne16end}              & $ 42.4 $ \\
        OCDC \cite{bojanowski14weakly}      & $ 43.9 $ \\
        HMM/RNN \cite{richard2017weakly}    & $ 46.3 $ \\
        \midrule
        \textbf{NN-Viterbi}              & $ \mathbf{48.7} $ \\
        \bottomrule
    \end{tabularx}
    \caption{Comparison of our method to several state-of-the-art methods
             for the task of action alignment. Results are reported as a
             variant of the Jaccard Index (intersection over detection).}
    \label{tab:actionAlignment}
    \vspace{-0.3cm}
\end{table}

The task of action alignment has first been addressed by Bojanowski \etal~\cite{bojanowski14weakly}.
In contrast to the previous task, the ordered action sequences $ \mathbf{c}_1^N $
are now also given for inference. Thus, only the alignment of actions to frames,
or in other words, the lengths $ \mathbf{l}_1^N $ of each segment, need to be
inferred. The training procedure is exactly the same as before.

The results are shown in Table~\ref{tab:actionAlignment}. Our method
outperforms the current state-of-the art by $ +2.4\% $.

\section{Conclusion}
\label{sec:conclusion}

We have proposed a direct learning algorithm that can handle weakly labeled
video sequences. The algorithm is generic and can be applied to any kind of model
whose best segmentation can be inferred using a Viterbi-like algorithm.
Unlike the CTC criterion, our approach allows to include multiple
explicitly modeled terms such as a context model and a length model, what has
been proven crucial for good performance.
Moreover, we showed that using an explicit length model and optimizing the
video classes directly leads to a huge improvement over related HMM-based
methods that use a pseudo ground-truth.
Overall, our method outperforms the current state-of-the-art by a large margin
and shows a robust and stable convergence behaviour.

\textbf{Acknowledgements.}
The work has been financially supported by the DFG projects KU 3396/2-1 (Hierarchical Models for Action Recognition and Analysis in Video Data) and
GA 1927/4-1 (DFG Research Unit FOR 2535 Anticipating Human Behavior) and the ERC Starting Grant ARCA (677650).
This work was supported by the AWS Cloud Credits for Research program and a Microsoft Azure Sponsorship.


{\small
\bibliographystyle{ieee}
\bibliography{references}

\begin{thebibliography}{10}\itemsep=-1pt

\bibitem{bojanowski14weakly}
P.~Bojanowski, R.~Lajugie, F.~Bach, I.~Laptev, J.~Ponce, C.~Schmid, and
  J.~Sivic.
\newblock Weakly supervised action labeling in videos under ordering
  constraints.
\newblock In {\em European Conf. on Computer Vision}, 2014.

\bibitem{bottou1998online}
L.~Bottou.
\newblock Online learning and stochastic approximations.
\newblock In {\em Online learning and neural networks}. Cambridge University
  Press, 1998.

\bibitem{bourlard2012connectionist}
H.~A. Bourlard and N.~Morgan.
\newblock {\em Connectionist speech recognition: a hybrid approach}, volume
  247.
\newblock Springer Science \& Business Media, 2012.

\bibitem{carreira2017quovadis}
J.~Carreira and A.~Zisserman.
\newblock Quo vadis, action recognition? a new model and the kinetics dataset.
\newblock In {\em IEEE Conf. on Computer Vision and Pattern Recognition}, 2017.

\bibitem{duchenne09automatic}
O.~Duchenne, I.~Laptev, J.~Sivic, F.~Bach, and J.~Ponce.
\newblock Automatic annotation of human actions in video.
\newblock In {\em Int. Conf. on Computer Vision}, 2009.

\bibitem{eyjolfsdottir2014detecting}
E.~Eyjolfsdottir, S.~Branson, X.~P. Burgos-Artizzu, E.~D. Hoopfer, J.~Schor,
  D.~J. Anderson, and P.~Perona.
\newblock Detecting social actions of fruit flies.
\newblock In {\em European Conf. on Computer Vision}, pages 772--787, 2014.

\bibitem{feichtenhofer2017residual}
C.~Feichtenhofer, A.~Pinz, and R.~P. Wildes.
\newblock Temporal residual networks for dynamic scene recognition.
\newblock In {\em IEEE Conf. on Computer Vision and Pattern Recognition}, 2017.

\bibitem{gan2016webly}
C.~Gan, C.~Sun, L.~Duan, and B.~Gong.
\newblock Webly-supervised video recognition by mutually voting for relevant
  web images and web video frames.
\newblock In {\em European Conf. on Computer Vision}, 2016.

\bibitem{gan2016youlead}
C.~Gan, T.~Yao, K.~Yang, Y.~Yang, and T.~Mei.
\newblock You lead, we exceed: Labor-free video concept learning by jointly
  exploiting web videos and images.
\newblock In {\em IEEE Conf. on Computer Vision and Pattern Recognition}, pages
  923--932, 2016.

\bibitem{graves2006connectionist}
A.~Graves, S.~Fern{\'a}ndez, F.~Gomez, and J.~Schmidhuber.
\newblock Connectionist temporal classification: labelling unsegmented sequence
  data with recurrent neural networks.
\newblock In {\em Int. Conf. on Machine Learning}, pages 369--376, 2006.

\bibitem{huang2016connectionist}
D.-A. Huang, L.~Fei-Fei, and J.~C. Niebles.
\newblock Connectionist temporal modeling for weakly supervised action
  labeling.
\newblock In {\em European Conf. on Computer Vision}, 2016.

\bibitem{koller2016deephand}
O.~Koller, H.~Ney, and R.~Bowden.
\newblock Deep hand: How to train a {CNN} on 1 million hand images when your
  data is continuous and weakly labelled.
\newblock In {\em IEEE Conf. on Computer Vision and Pattern Recognition}, 2016.

\bibitem{koller2017resign}
O.~Koller, S.~Zargaran, and H.~Ney.
\newblock Re-sign: Re-aligned end-to-end sequence modelling with deep recurrent
  {CNN}-{HMM}s.
\newblock In {\em IEEE Conf. on Computer Vision and Pattern Recognition}, 2017.

\bibitem{kuehne14language}
H.~Kuehne, A.~B. Arslan, and T.~Serre.
\newblock The language of actions: Recovering the syntax and semantics of
  goal-directed human activities.
\newblock In {\em IEEE Conf. on Computer Vision and Pattern Recognition}, 2014.

\bibitem{kuehne16end}
H.~Kuehne, J.~Gall, and T.~Serre.
\newblock An end-to-end generative framework for video segmentation and
  recognition.
\newblock In {\em IEEE Winter Conf. on Applications of Computer Vision}, 2016.

\bibitem{kuehne2017weakly}
H.~Kuehne, A.~Richard, and J.~Gall.
\newblock Weakly supervised learning of actions from transcripts.
\newblock {\em Computer Vision and Image Understanding}, 2017.

\bibitem{laptev08learning}
I.~Laptev, M.~Marszalek, C.~Schmid, and B.~Rozenfeld.
\newblock Learning realistic human actions from movies.
\newblock In {\em IEEE Conf. on Computer Vision and Pattern Recognition}, 2008.

\bibitem{lea2017temporal}
C.~Lea, M.~D. Flynn, R.~Vidal, A.~Reiter, and G.~D. Hager.
\newblock Temporal convolutional networks for action segmentation and
  detection.
\newblock In {\em IEEE Conf. on Computer Vision and Pattern Recognition}, 2017.

\bibitem{lea2016segmental}
C.~Lea, A.~Reiter, R.~Vidal, and G.~D. Hager.
\newblock Segmental spatiotemporal {CNN}s for fine-grained action segmentation.
\newblock In {\em European Conf. on Computer Vision}, pages 36--52, 2016.

\bibitem{lin2017ctc}
M.~Lin, N.~Inoue, and K.~Shinoda.
\newblock {CTC} network with statistical language modeling for action sequence
  recognition in videos.
\newblock In {\em Proceedings of the Thematic Workshops of the ACM Conf. on
  Multimedia}, pages 393--401, 2017.

\bibitem{marszalek09actions}
M.~Marszalek, I.~Laptev, and C.~Schmid.
\newblock Actions in context.
\newblock In {\em IEEE Conf. on Computer Vision and Pattern Recognition}, 2009.

\bibitem{ni2014multiple}
B.~Ni, V.~R. Paramathayalan, and P.~Moulin.
\newblock Multiple granularity analysis for fine-grained action detection.
\newblock In {\em IEEE Conf. on Computer Vision and Pattern Recognition}, pages
  756--763, 2014.

\bibitem{richard2016temporal}
A.~Richard and J.~Gall.
\newblock Temporal action detection using a statistical language model.
\newblock In {\em IEEE Conf. on Computer Vision and Pattern Recognition}, 2016.

\bibitem{richard2017weakly}
A.~Richard, H.~Kuehne, and J.~Gall.
\newblock Weakly supervised action learning with {RNN} based fine-to-coarse
  modeling.
\newblock In {\em IEEE Conf. on Computer Vision and Pattern Recognition}, 2017.

\bibitem{richard2018actionsets}
A.~Richard, H.~Kuehne, and J.~Gall.
\newblock Weakly supervised action segmentation without ordering constraints.
\newblock In {\em IEEE Conf. on Computer Vision and Pattern Recognition}, 2018.

\bibitem{rohrbach2012database}
M.~Rohrbach, S.~Amin, M.~Andriluka, and B.~Schiele.
\newblock A database for fine grained activity detection of cooking activities.
\newblock In {\em IEEE Conf. on Computer Vision and Pattern Recognition}, pages
  1194--1201, 2012.

\bibitem{simonyan2014two}
K.~Simonyan and A.~Zisserman.
\newblock Two-stream convolutional networks for action recognition in videos.
\newblock In {\em Advances in Neural Information Processing Systems}, pages
  568--576, 2014.

\bibitem{singh16multistream}
B.~Singh, T.~K. Marks, M.~Jones, O.~Tuzel, and M.~Shao.
\newblock A multi-stream bi-directional recurrent neural network for
  fine-grained action detection.
\newblock In {\em IEEE Conf. on Computer Vision and Pattern Recognition}, 2016.

\bibitem{stein2013combining}
S.~Stein and S.~J. McKenna.
\newblock Combining embedded accelerometers with computer vision for
  recognizing food preparation activities.
\newblock In {\em ACM Int. Joint Conf. on Pervasive and Ubiquitous Computing},
  pages 729--738, 2013.

\bibitem{sun15temporal}
C.~Sun, S.~Shetty, R.~Sukthankar, and R.~Nevatia.
\newblock Temporal localization of fine-grained actions in videos by domain
  transfer from web images.
\newblock In {\em ACM Conf. on Multimedia}, 2015.

\bibitem{tang2012learning}
K.~Tang, L.~Fei-Fei, and D.~Koller.
\newblock Learning latent temporal structure for complex event detection.
\newblock In {\em IEEE Conf. on Computer Vision and Pattern Recognition}, pages
  1250--1257, 2012.

\bibitem{vo2014stochastic}
N.~N. Vo and A.~F. Bobick.
\newblock From stochastic grammar to bayes network: Probabilistic parsing of
  complex activity.
\newblock In {\em IEEE Conf. on Computer Vision and Pattern Recognition}, pages
  2641--2648, 2014.

\bibitem{wang2013action}
H.~Wang and C.~Schmid.
\newblock Action recognition with improved trajectories.
\newblock In {\em Int. Conf. on Computer Vision}, pages 3551--3558, 2013.

\bibitem{wang2017untrimmed}
L.~Wang, Y.~Xiong, D.~Lin, and L.~Van~Gool.
\newblock Untrimmednets for weakly supervised action recognition and detection.
\newblock In {\em IEEE Conf. on Computer Vision and Pattern Recognition}, 2017.

\bibitem{wang2016temporal}
L.~Wang, Y.~Xiong, Z.~Wang, Y.~Qiao, D.~Lin, X.~Tang, and L.~Van~Gool.
\newblock Temporal segment networks: Towards good practices for deep action
  recognition.
\newblock In {\em European Conf. on Computer Vision}, pages 20--36, 2016.

\bibitem{yeung2016endtoend}
S.~Yeung, O.~Russakovsky, G.~Mori, and L.~Fei-Fei.
\newblock End-to-end learning of action detection from frame glimpses in
  videos.
\newblock In {\em IEEE Conf. on Computer Vision and Pattern Recognition}, 2016.

\bibitem{zhao2017segment}
Y.~Zhao, Y.~Xiong, L.~Wang, Z.~Wu, X.~Tang, and D.~Lin.
\newblock Temporal action detection with structured segment networks.
\newblock In {\em Int. Conf. on Computer Vision}, 2017.

\end{thebibliography}
}

\end{document}